\documentclass[runningheads]{llncs}
\usepackage{graphicx}
\usepackage{epsfig}
\usepackage{amsmath}
\usepackage{amssymb} 
\usepackage{times}
\usepackage{color}
\usepackage{multirow}
\usepackage{caption}
\usepackage{float}
\usepackage{tabularx}
\usepackage{ragged2e}
\usepackage{hhline}
\usepackage{subcaption}
\usepackage{url}
\usepackage{hyperref}
\usepackage[symbol]{footmisc}

\newcommand{\cutabstractup}{\vspace*{-0.2in}}
\newcommand{\cutabstractdown}{\vspace*{-0.2in}}
\newcommand{\cutsectionup}{\vspace*{-0.2in}}
\newcommand{\cutsectiondown}{\vspace*{-0.12in}}
\newcommand{\cutsubsectionup}{\vspace*{-0.1in}}
\newcommand{\cutsubsectiondown}{\vspace*{-0.07in}}

\begin{document}
\title{MT-VAE: Learning Motion Transformations \\ to Generate Multimodal Human Dynamics}

\titlerunning{Learning Motion Transformations to Generate Multimodal Human Dynamics} 
%

\author{
  Xinchen Yan$^{1*}$ \quad Akash Rastogi$^1$ \quad Ruben Villegas$^1$ \quad Kalyan Sunkavalli$^2$\\ 
  Eli Shechtman$^2$ \quad
  Sunil Hadap$^2$ \quad Ersin Yumer$^3$ \quad Honglak Lee$^{1,4}$\\
}

\institute{University of Michigan, Ann Arbor
\and 
Adobe Research
\and
Argo AI
\and
Google Brain}

\authorrunning{X. Yan et al.}
%

\maketitle              

\begin{abstract}
\cutabstractup
Long-term human motion can be represented as a series of \emph{motion modes}---motion sequences that capture short-term temporal dynamics---with transitions between them.
We leverage this structure and present a novel \emph{Motion Transformation Variational Auto-Encoders (MT-VAE)} for learning motion sequence generation.
Our model jointly learns a feature embedding for motion modes (that the motion sequence can be reconstructed from) and a feature transformation that represents the transition of one motion mode to the next motion mode.
Our model is able to generate multiple diverse and plausible motion sequences in the future from the same input.
We apply our approach to both facial and full body motion, and demonstrate applications like analogy-based motion transfer and video synthesis.
\cutabstractdown
\end{abstract}

\cutsectionup
\section{Introduction}
\cutsectiondown
\footnotetext[1]{Work partially done during internship with Adobe Research.}
Modeling the dynamics of human motion --- both facial and full body motion --- is a fundamental problem in computer vision, graphics, and machine intelligence, with applications ranging from virtual characters~\cite{deAguiar2008performance,beeler2011performance}, video-based animation and editing~\cite{yang2011expression,suwajanakorn2015makes,suwajanakorn2017synthesizing}, and human-robot interfaces~\cite{sermanet2017time}.    
Human motion is known to be highly structured and can be modeled as a sequence of atomic units that we refer to as \emph{motion modes}.
A motion mode captures the \emph{short-term} temporal dynamics of a human action (e.g., smiling or walking), including its related stylistic attributes (e.g., how wide is the smile, how fast is the walk). 
Over the \emph{long-term}, a human action sequence can be segmented into a series of motion modes with transitions between them (e.g., a transition from a neutral expression to smiling to laughing).
This structure is well known (referred to as basis motions~\cite{rose1996motiontransitions} or walk cycles) and widely used in computer animation.

\begin{figure}[t]
\centering
\includegraphics[width=0.75\linewidth]{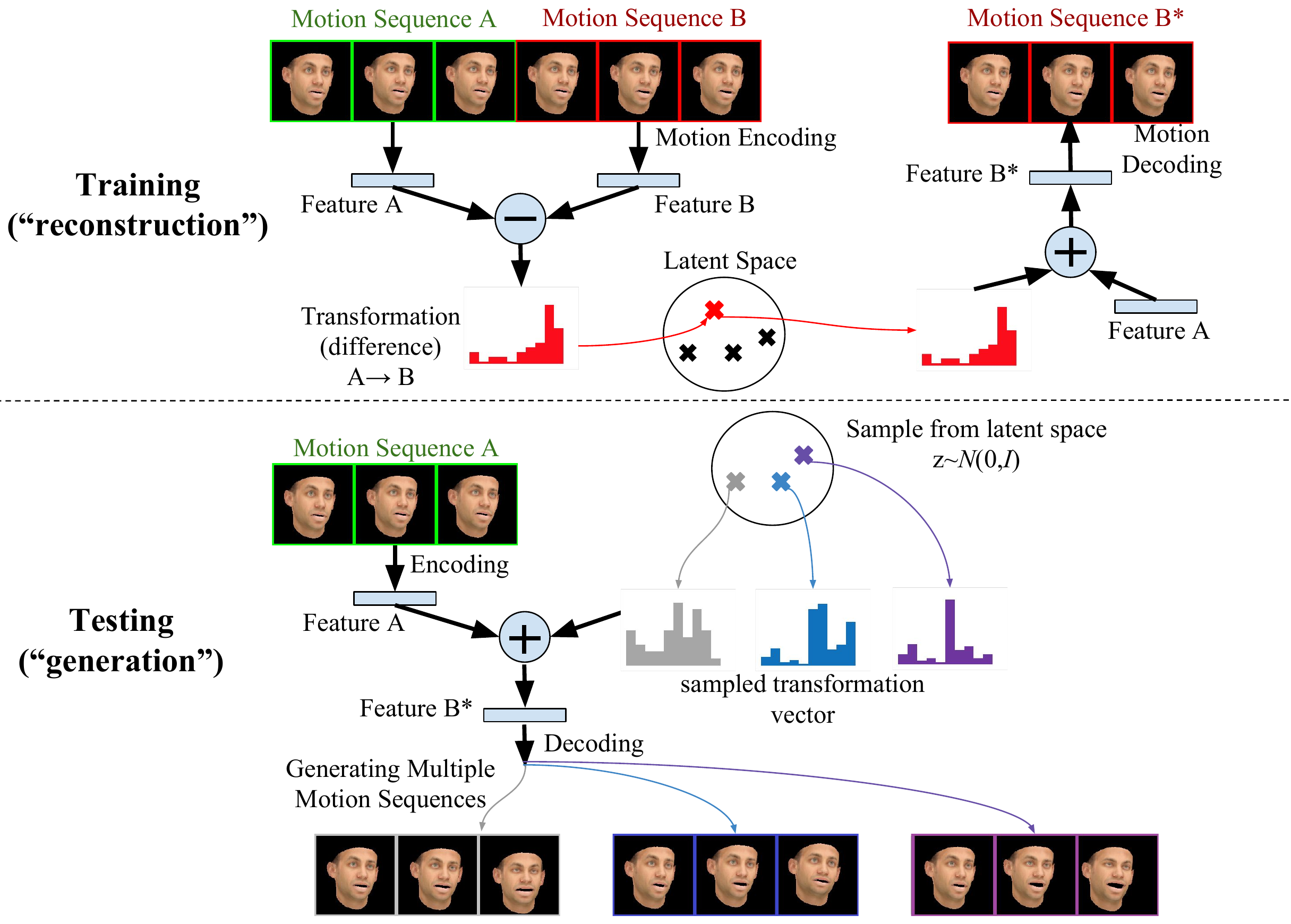}
\caption{Top: Learning motion sequence generation using \emph{Motion Transformation VAE}. 
Bottom: Generating multiple future motion sequences from the transformation space.}
\label{figure:figure_intro}
\vspace*{-0.25in}
\end{figure}

This paper leverages this structure to learn to generate human motion sequences, i.e., given a short human action sequence (present motion mode), we want to synthesize the action going forward (future motion mode).
We hypothesize that (1) each motion mode can be represented as a low-dimensional feature vector, and (2) transitions between motion modes can be modeled as \textit{transformations} of these features. 
As shown in Figure~\ref{figure:figure_intro}, we present a novel model termed \emph{Motion Transformation Variational Auto-Encoders (MT-VAE)} for learning motion sequence generation.
Our MT-VAE is implemented using an LSTM encoder-decoder that embeds each short sub-sequence into a feature vector that can be decoded to reconstruct the motion.
We further assume that the transition between current and future modes can be captured by a certain transformation.
In the paper, we demonstrate that the proposed MT-VAE learns a motion feature representation in an unsupervised way. 

A challenge with human motion is that it is inherently multimodal, i.e., the same initial motion mode could transition into different motion modes (e.g., a smile could transition to a frown, or a smile while looking left, or a wider smile, etc.).
A deterministic model would not be able to learn these variations and may collapse to a single-mode distribution.
Our MT-VAE supports a stochastic sampling of the feature transformations to generate multiple plausible output motion modes from a single input.
This allows us to model transitions that may be rare (or potentially absent) in the the training set.

We demonstrate our approach on both facial and full human body motions. 
In both domains, we conduct extensive ablation studies and comparisons with previous work showing that our generation results are more plausible (i.e., better preserve the structure of human dynamics) and diverse (i.e., explore multiple motion modes).
We further demonstrate applications like 
1) analogy-based motion transfer
(e.g., transferring the act of smiling from one pose to another pose)
and 
2) future video synthesis (i.e., generating multiple possible future videos given input frames with human motions). 
Our key contributions are summarized as follows:
\vspace{-0.05in}
\begin{itemize}
    \item We propose a generative motion model that consists of a sequence-level motion feature embedding and feature transformations, and show that it can be trained in an unsupervised manner.
    \item We show that stochastically sampling the transformation space is able to generate future motion dynamics that are diverse and plausible.
    \item We demonstrate applications of the learned model to challenging tasks like motion transfer and future video synthesis for both facial and human body motions.
\vspace*{-0.1in}
\end{itemize}

\cutsectionup
\section{Related Work}
\cutsectiondown
Understanding and modeling human motion dynamics has been a long-standing problem for decades \cite{bregler1997learning,efros2003recognizing,gorelick2007actions}.
Due to the high dimensionality of video data, early work mainly focused on learning hierarchical spatio-temporal representations for video event and action recognition~\cite{laptev2005space,wang2011action,wang2012mining}.
In recent years, predicting and synthesizing motion dynamics using deep neural networks has become a popular research topic.
Walker et al.~\cite{walker2015dense}, Fischer et al.~\cite{fischer2015flownet} learn to synthesize dense flow in the future from a single image.
Walker et al.~\cite{walker2016uncertain} extended the deterministic prediction framework by modeling the flow uncertainty using variational auto-encoders.
Chao et al.~\cite{chao2017forecasting} proposed a recurrent neural network to generate movement of 3D human joints from a single observation with a 3D in-network projection layer.
Taking one step further,  Villegas et al.~\cite{villegas2017learning}, Walker et al.~\cite{walker2017pose} explored hierarchical structure (e.g., 2D human joints) for motion prediction in the future using recurrent neural networks.
Li et al.~\cite{li2017auto} proposed an auto-conditional recurrent framework to generate long-term human motion dynamics through time.
Besides human motion, face synthesis and editing is another interesting topic in vision and graphics.
Methods for reenacting and interpolating face sequences in video have been developed \cite{yang2011expression,yang2012facial,thies2016face2face,elor2017bringingPortraits} based on a 3D morphable face representation~\cite{blanz1999morphable}.
Very recently, Suwajanakorn et al.~\cite{suwajanakorn2017synthesizing} introduced a speech-driven face synthesis system that learns to generate lip motions with a recurrent neural network.

Besides the flow representation, motion synthesis has been explored in a broader context, namely, video generation.
For example, synthesizing video sequence in the future from a single or multiple video frames as initialization.
Early works employed patch-based method for short-term video generation using mean squared mean squared loss~\cite{srivastava2015unsupervised} or perceptual loss~\cite{mathieu2015deep}.
Given an atomic action as additional condition, previous works extended with action-conditioned (i.e., rotation, location, etc) architectures that enable better semantic control in video generation~\cite{hinton2011transforming,Oh15,FinnGL16,yang2015weakly}.
Due to the difficulty in holistic video frame prediction, the idea of disentangling video factors into motion and content is explored in \cite{villegas2017decomposing,denton2017unsupervised,xue2016visual,vondrick2016generating,tulyakov2017mocogan,nevan2018hierarchical}.
Video generation has also been approached with architectures that output multinomial distribution vectors over the possible pixel values for each pixel in the generated frame~\cite{kalchbrenner2016video}.

The notion of feature transformations has also been exploited for other tasks. Mikolov et al.~\cite{mikolov2013distributed} showcased the composition additive property of word vectors learned in an unsupervised way from language data;
Kulkarni et al.~\cite{kulkarni2015deep}, Reed et al.~\cite{reed2015deep} suggested that additive transformation can be achieved via reconstruction or prediction task by learning from parallel paired image data.
In the video domain, Wang et al.~\cite{wang2016actions} studied a transformation-aware representation for semantic human action classification;
Zhou et al.~\cite{zhou2016learning} investigated time-lapse video generation given additional class labels.

Multimodal conditional generation has recently been explored for images~\cite{sohn2015learning,zhu2017toward}, sketch drawings~\cite{ha2017neural}, natural language \cite{bowman2015generating,hu2017controllable}, and video prediction~\cite{mohammad2018stochastic,denton2018stochastic}.
As noted in previous work, learning to generate diverse and plausible visual data is very challenging for the following reasons:
first, mode collapse may occur without one-to-many pairs. Collecting sequence data where one-to-many pairs exist is non-trivial.
Second, posterior collapse could happen when the generation model is based on a recurrent neural network.

\cutsectionup
\section{Problem Formulation and Methods}
\cutsectiondown
We start by giving an overview of our problem.
We are given a sequence of $T$ observations $S_A = [x_1, x_2, \cdots, x_T]$, where $x_t \in \mathbb{R}^{D}$ is a $D$ dimensional vector representing the observation at time $t$.
These observations encode the structure of the moving object and can be represented in different ways, for e.g., as keypoint locations or shape and pose parameters. 
\emph{Changes} in these observations encode the motion that we are interested in modeling.
We refer to the entire sequence as a \emph{motion mode}.
Given a motion mode, $S_A \in \mathbb{R}^{T \times D}$, we aim to build a model that is capable of predicting a future motion mode, $S_B = [y_1, y_2, \cdots, y_T] $, where $y_t \in \mathbb{R}^{D}$ represents the predicted $t$-th step in the \emph{future}, i.e., $y_1 = x_{T+1}$. 
%
%
We first start with a discussion of two potential baseline models that could be used for this task (Section~\ref{sec:preliminaries}), and then present our method (Section~\ref{sec:tcvae_transformation}).

\cutsubsectionup
\subsection{Preliminaries}
\label{sec:preliminaries}
\cutsubsectiondown

\begin{figure*}[t]
\centering
\includegraphics[width=0.99\linewidth]{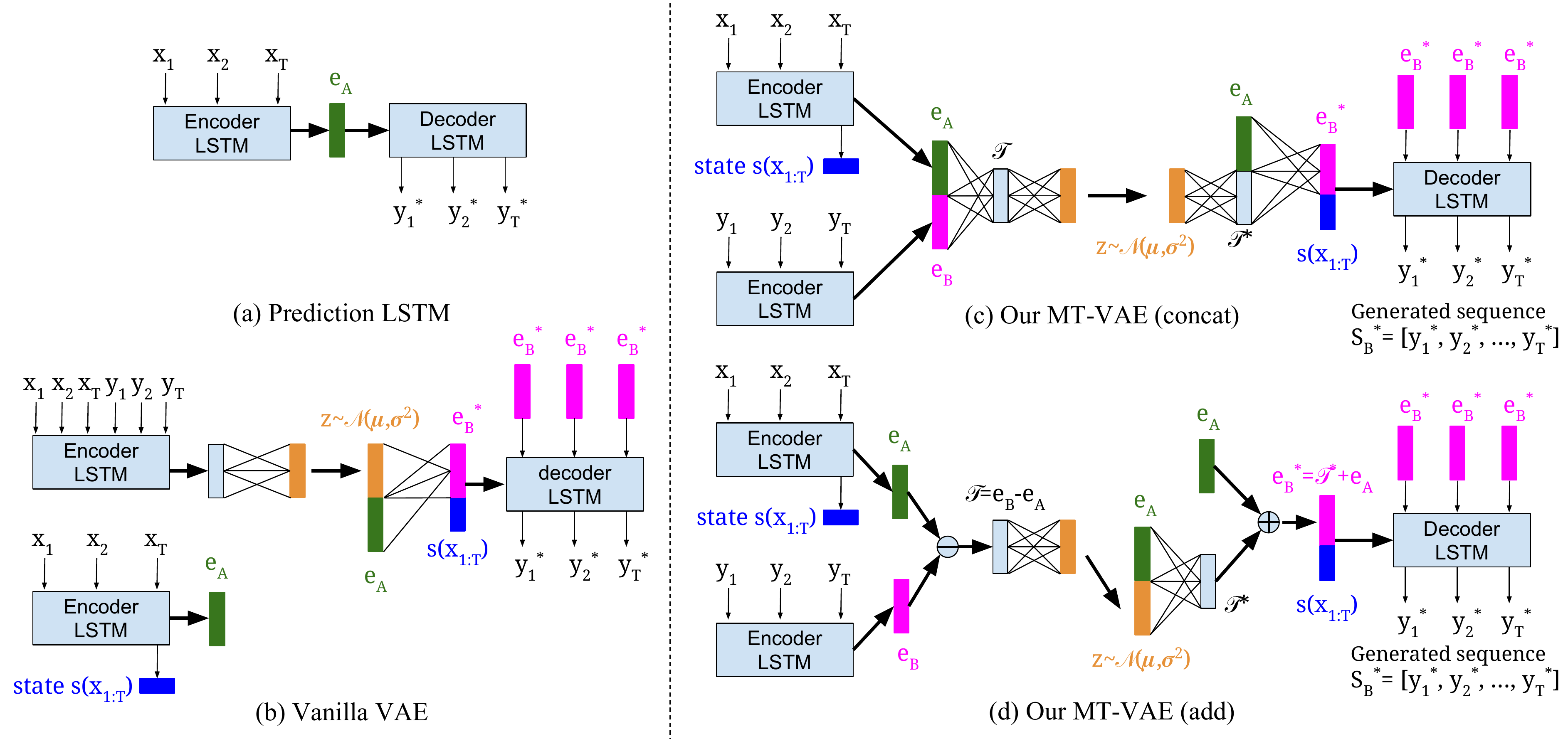}
\caption{Illustrations of different models for motion sequence generation. $s(x_{1:T})$ indicates the hidden state of the Encoder LSTM at time $T$. 
}
\label{figure:figure_arch}
\vspace*{-0.2in}
\end{figure*}

\paragraph{Prediction LSTM for Sequence Generation.}
Figure~\ref{figure:figure_arch}(a) shows a simple encoder-decoder LSTM \cite{hochreiter1997long,srivastava2015unsupervised} as a baseline for the motion prediction task.
At time $t$, the encoder LSTM takes the motion $x_t$ as input and updates its internal representation.
After going through the entire motion mode $S_A$, it outputs a fixed-length feature $e_A \in \mathbb{R}^{N_e}$ as an intermediate representation.
We initialize the internal representation of decoder LSTM using the feature $e_A$ computed.
At time $t$ of the decoding stage, the decoder LSTM predicts the motion $y_t$.
This way, the decoder LSTM gradually predicts the entire motion mode $S^*_B = [y_1, y_2, \cdots, y_T]$ in the future within $T$ steps.
We denote the encoder LSTM as function $f: \mathbb{R}^{T\times D} \rightarrow \mathbb{R}^{N_e}$ and the decoder LSTM as function $g: \mathbb{R}^{N_e} \rightarrow \mathbb{R}^{T\times D}$.
As a design choice, we initialize the decoder LSTM with additional input $x_T$ for smoother prediction.

\vspace*{-0.1in}
\paragraph{Vanilla VAE for Sequence Generation.}
As the deterministic LSTM model fails to reflect the multimodal nature of human motion, we consider a statistical model, $p_\theta(S_B|S_A)$, parameterized by $\theta$.
Given the observed sequence $S_A$, the model estimates a probability for the possible future sequence $S_B$ instead of a single outcome.
To model the multimodality (i.e., $S_A$ can transition to different $S_B$'s), a latent variable $z$ (sampled from prior distribution) is introduced to capture the inherent uncertainty.
The future sequence $S_B$ is generated as follows:
%
\begin{enumerate}
\item Sample latent variable $z \sim \mathcal{N}(\mathbf{0}, \mathbf{I})$;
\item Given $S_A$ and $z$, generate a sequence of length $T$: $S_B\sim p_\theta(S_B|z, S_A)$; 
\vspace*{-0.2cm}
\end{enumerate}

Following previous work on VAEs~\cite{kingma2013auto,sohn2015learning,gregor2015draw,yan2016attribute2image,walker2016uncertain,xue2016visual,walker2017pose},
the objective is to maximize the variational lower-bound of the conditional log-probability $\log p_\theta(S_B|S_A)$:
\begin{align}
\mathcal{L}_\text{VAE} = -\textit{KL}(q_\phi (z|S_B,S_A) || p_\theta (z)) + \mathbb{E}_{q_\phi (z|S_B,S_A)} \big[ \log p_\theta (S_B|S_A,z) \big]\label{eqn:method-lb2}
\end{align}
In Eq.~\ref{eqn:method-lb2}, $q_\phi(z|S_B, S_A)$ is referred as an auxiliary posterior that approximates the true posterior $p_\theta(z|S_B, S_A)$.
Specifically, the prior $p_\theta(z)$ is assumed to be $\mathcal{N}(\mathbf{0}, \mathbf{I})$.
The posterior $q_\phi(z|S_B, S_A)$ is a multivariate Gaussian distribution with mean and variance $\mu_\phi$ and $\sigma_{\phi}^2$, respectively.
Intuitively, the first term in Eq.~\ref{eqn:method-lb2} regularizes the auxiliary posterior $q_\phi(z|S_B,S_A)$ with prior $p_\theta(z)$.
The second term $\log p_\theta (S_B|S_A,z)$ can be considered as an auto-encoding loss, where we refer to $q_\phi(z|S_B, S_A)$ as an encoder or recognition model, and $p_\theta(S_B|z, S_A)$ as a decoder or generation model.

As shown in Figure~\ref{figure:figure_arch}(b), the vanilla VAE model adopts similar LSTM encoder and decoder for sequence processing. 
In contrast to Prediction LSTM model, the vanilla VAE decoder takes both motion feature $e_A$ and latent variable $z$ into account.
Ideally, this allows to generate diverse motion sequences by drawing different samples from the latent space.
However, the semantic role of the latent variable $z$ in this vanilla VAE model is not straight-forward and may not effectively represent long-term trends (e.g., dynamics in a specific motion mode or during change of modes).

\cutsubsectionup
\subsection{Motion-to-Motion Transformations in Latent Space}
\label{sec:tcvae_transformation}
\cutsubsectiondown

To further improve motion sequence generation beyond vanilla VAE, we propose to explicitly enforce the structure of motion modes in the latent space.
We assume that (1) each motion mode can be represented as low-dimensional feature vector, and (2) transitions between motion modes can be modeled as \textit{transformations} of these features.
Our design is also supported by early studies on hierarchical motion modeling and prediction \cite{bregler1997learning,smith2013sources,lan2014hierarchical}.

We present a \emph{Motion Transformation VAE (or MT-VAE)} (Fig.~\ref{figure:figure_arch}(c)) with four components:
\begin{enumerate}
\item An LSTM encoder $f: \mathbb{R}^{T\times D} \rightarrow \mathbb{R}^{N_e}$ maps the input sequences into motion features through $e_A = f(S_A)$ and $e_B = f(S_B)$, respectively.
\item A latent encoder $h_{e\rightarrow z}: \mathbb{R}^{2 \times N_e} \rightarrow \mathbb{R}^{N_z}$ computes the transformation in the latent space $z = h_{e\rightarrow z}([e_A, e_B])$ by concatenating motion features $e_A$ and $e_B$.
Here, $N_z$ indicates the latent space dimension.
\item A latent decoder $h_{z\rightarrow e}: \mathbb{R}^{N_z + N_e} \rightarrow \mathbb{R}^{N_e}$ synthesizes the motion feature in the future from latent transformation $z$ and current motion feature $e_A$ via $e^*_B = h_{z\rightarrow e}([z, e_A])$.
\item An LSTM decoder $g: \mathbb{R}^{N_e} \rightarrow \mathbb{R}^{T\times D}$ synthesizes the future sequence given motion feature: $S^*_B = g(e^*_B)$.
\end{enumerate}
Similar to the Prediction LSTM, we use an LSTM encoder/decoder to map motion modes into feature space. The MT-VAE further maps these features into \emph{latent transformations} and stochastically samples these transformations. As we demonstrate, this change makes the model more expressive and leads to more plausible results. Finally, in the sequence decoding stage of MT-VAE, we feed the synthesized motion feature $e^*_B$ as input to the decoder LSTM, with internal state initialized using the same motion feature $e^*_B$ with an additional input $x_t$.

\cutsubsectionup
\subsection{Additive Transformations in Latent Space}
\label{sec:tcvae_learning}
\cutsubsectiondown

Although MT-VAE explicitly models motion transformations in latent space, this space might be unconstrained because the transformations are computed from vector concatenation of motion features $e_A$ and $e_B$ in our latent encoder $h_{e\rightarrow z}$.
To better regularize the transformation space, we present an additive variant of MT-VAE, that is depicted in Figure~\ref{figure:figure_arch}(d).
To distinguish between the two variants, we call the previous model \emph{MT-VAE (concat)} and this model \emph{MT-VAE (add)}, respectively.
Our model is inspired by recent success of \emph{deep analogy-making} methods~\cite{reed2015deep,villegas2017decomposing} where a relation (or transformation) between two examples can be represented as a difference in the embedding space.
In this model, we strictly constrain the latent encoding and decoding steps as follows:
%
\begin{enumerate}
    \item Our latent encoder $h_{\mathcal{T}\rightarrow z}: \mathbb{R}^{N_e} \rightarrow \mathbb{R}^{N_z}$ computes the difference between two motion features $e_A$ and $e_B$ via $\mathcal{T} = e_B - e_A$ ; then it maps the difference feature $\mathcal{T}$ into a transformation in the latent space via $z = h_{\mathcal{T}\rightarrow z}(\mathcal{T})$.
    \item Our latent decoder $h_{z\rightarrow \mathcal{T}}: \mathbb{R}^{N_z + N_e} \rightarrow \mathbb{R}^{N_e}$ reconstructs the difference feature $\mathcal{T}^*$ from latent variable $z$ and current motion feature $e_A$ via $\mathcal{T}^* = h_{z\rightarrow \mathcal{T}}(z, e_A)$.
    \item Finally, we apply a simple additive interaction to reconstruct the motion feature via $e^*_B = e_A + \mathcal{T}^*$;\\
\end{enumerate}
\vspace*{-0.16in}

In step one, we infer the latent variable using $h_{\mathcal{T}\rightarrow z}$ from the difference of $e_A$ and $e_B$ (instead of a applying a linear layer on concatenated vectors).
Intuitively, the latent code is expected to capture the mode transition from the current motion to the future motion rather than a concatenation of two modes.
In step two, we reconstruct the transformation from the latent variable via $h_{z\rightarrow \mathcal{T}}(z, e_A)$ where $z$ is obtained from recognition model.
In this design, the feature difference is dependent on both latent transformation $z$ and current motion feature $e_A$.
Alternatively, we can make our latent decoder $h_{z\rightarrow \mathcal{T}}$ context-free by removing input from motion feature $e_A$.
This way, the latent decoder is supposed to hallucinate the motion difference solely from the latent space. We provide this ablation study in Section~\ref{sec:exp_seqgen}.

\begin{figure*}[t]
\centering
\includegraphics[width=0.99\linewidth]{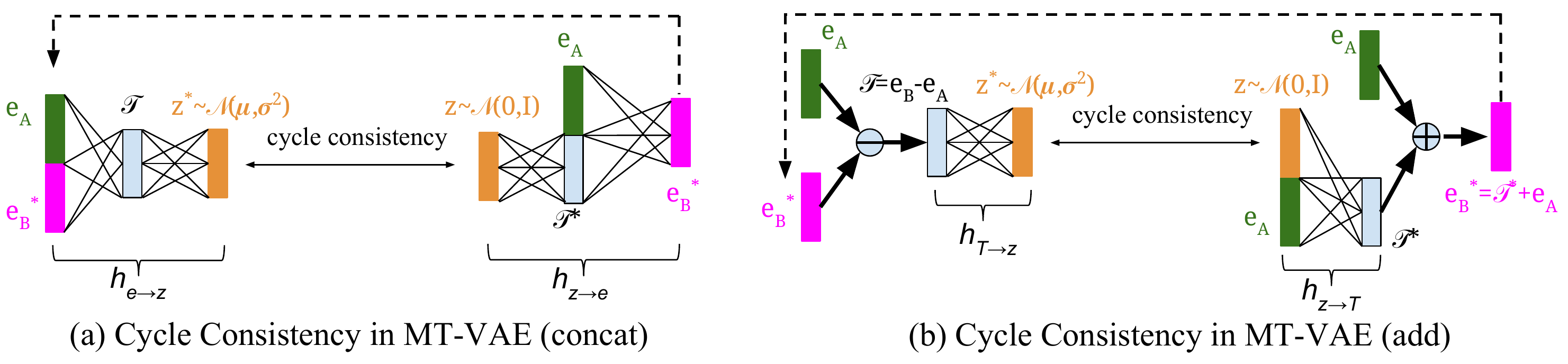}
\caption{Illustrations of cycle consistency in MT-VAE variations. }
\label{figure:figure_cycle}
\vspace*{-0.2in}
\end{figure*}

Besides the architecture-wise regularization, we introduce two additional objectives while training our model.
\vspace*{-0.1in}
\paragraph{Cycle Consistency.} 
As mentioned previously, our training objective $\mathcal{L}_\text{VAE}$ in Eq.~\ref{eqn:method-lb2} is composed of a KL term and a reconstruction term at each frame.
The KL term regularizes the latent space, while the reconstruction term ensures that the data can be explained by our generative model.
However, we do not have direct regularization in the feature space.
We therefore introduce a cycle-consistency loss in Eq.~\ref{eqn:method-cycle} (for MT-VAE (concat)) and Eq.~\ref{eqn:method-cycle-add} (for MT-VAE (add)).
Figure~\ref{figure:figure_cycle} illustrates the cycle consistency in details.
\begin{align}
\mathcal{L}_\text{cycle}^\text{concat} &= ||z^* - z||\text{, where $z^* = h_{e \rightarrow z}([e_A, h_{z\rightarrow e}(z, e_A)])$ and $z \sim \mathcal{N}(\mathbf{0}, \mathbf{I})$}\label{eqn:method-cycle}\\
\mathcal{L}^\text{add}_\text{cycle} &= ||z^* - z||\text{, where $z^* = h_{\mathcal{T} \rightarrow z}(h_{z\rightarrow \mathcal{T}}(z, e_A))$ and $z \sim \mathcal{N}(\mathbf{0}, \mathbf{I})$}
\label{eqn:method-cycle-add}
\end{align}
In our preliminary experiments, we also investigated a consistency loss with a bigger cycle (involving the actual motion sequences) during training but we found it ineffective as a regularization term in our setting.
We hypothesize that vanishing or exploding gradients make the cycle-consistency objective less effective, which is a known issue when training recurrent neural networks.

\vspace*{-0.1in}
\paragraph{Motion Coherence.}

Specific to our motion generation task, we introduce a motion coherence loss in Eq.~\ref{eqn:method-motion} that encourages a smooth transition in velocity in the first $K$ steps of prediction. 
We define the velocity $v_1 = y_1 - x_T$ and $v_k = y_k - y_{k-1}$ when $k \geq 2$.
Intuitively, such loss prevents the generated sequence from deviating too far from the future sequence sampled from the prior.
\vspace*{-0.1in}
\begin{align}
\mathcal{L}_\text{motion} = \frac{1}{K}\sum_{t=1}^K ||v^*_t - v_t||\text{, where $g(e^z_B) = [y^*_1, \cdots, y^*_T]$ and $z \sim \mathcal{N}(\mathbf{0}, \mathbf{I}$)}\label{eqn:method-motion}
\end{align}

Finally, we summarize our overall loss in Eq.~\ref{eqn:method-full}, where $\lambda_\text{cycle}$ and $\lambda_\text{motion}$ are two balancing hyper-parameters for cycle consistency and motion coherence, respectively.
\begin{align}
\mathcal{L}_\text{MT-VAE} = \mathcal{L}_\text{VAE} + \lambda_\text{cycle}  \mathcal{L}_\text{cycle} + \lambda_\text{motion} \mathcal{L}_\text{motion}
\label{eqn:method-full}
\end{align}

\cutsectionup
\section{Experiments}
\cutsectiondown

\paragraph{Datasets.} The evaluation is conducted on the datasets involving two representative human motion modeling tasks: Affect-in-the-wild  (Aff-Wild)~\cite{zafeiriou2017aff} for facial motions and Human3.6M~\cite{ionescu2014human3} for full body motions.
The Aff-Wild dataset contains more than 400 video clips (2,000 minutes in total) collected from Youtube with natural facial expression and head motion patterns.
To better focus on face motion modeling (e.g., expressions and head movements), we leveraged the 3D morphable face model~\cite{bfm09,blanz1999morphable} (e.g., face identity, face expression, and pose) in our experiments.
We fitted 198-dim identity coefficients, 29-dim expression coefficients, and 6-dim pose parameters to each frame with a pre-trained 3DMM-CNN~\cite{tran2017regressing} model, followed by a face fitting algorithm~\cite{zhu2016face} based on optimization.
This disentangled representation allows us to study face motion modeling without being distracted by unrelated factors such as facial identity, background scene, and illumination of the environment.
We trained our model with 80\% of the data on the expression and pose parameters since these are the main factors that change over time.
Human3.6M is a large-scale database containing more than 800 human motion sequences captured by 11 professional actors (3.6 million frames in total) in an indoor environment.
For experiments on Human3.6M, we used the raw 2D trajectories of 32 keypoints and further normalized the data into coordinates within the range $[-1, 1]$.
We used subjects number 1, 5, 6, 7, and 8, for training and tested on subjects 9 and 11.
We used 5\% of the training data for the purpose of model validation.
\paragraph{Architecture Design.}
\vspace*{-0.1in}
Our MT-VAE model consists of four components: sequence encoder network, sequence decoder network, latent encoder network, and latent decoder network.
We build our sequence encoder and decoder using Long Short-term Memory units (LSTMs)~\cite{hochreiter1997long}. We used 1-layer LSTM with 1,024 hidden units for both networks.
For experiments on Aff-Wild dataset, the input to our sequence encoder is the 35-dimensional expression-pose representation (29 expression and 6 pose parameters) per timestep and we recursively predict the future parameters using our sequence decoder.
For experiments on Human3.6M dataset, we used the 64-dimensional xy-coordinate representation (32 joints with 2 coordinates each joint) instead.
Given past and future motion features extracted from our sequence encoder network, we build three fully-connected layers with skip connections within our latent encoding network.
We adopted a similar architecture (three fully-connected layers with skip connections) for our latent decoder network.
For all the models (including baselines), we fixed the bottleneck latent dimension to be 512 and found this configuration is sufficient to generate both face and full-body motions.

\paragraph{Implementation Details.}
We used ADAM~\cite{kingma2014adam} for optimization in all experiments.
For training, we used a mini-batch size of 256 and learning rate of 0.0001 with default ADAM settings (e.g., $\beta_1 = 0.9, \beta_2 = 0.999$).
For experiments on Aff-Wild, we trained models to predict 32 steps in the future given a varying number of observed frames between 8 and 16.
For experiments on Human3.6M, we trained models to predict 64 steps in the future given a varying number of observed frames between 10 and 20.
To stabilize the training, we applied layer normalization~\cite{ba2016layer} in both LSTMs and fully-connected layers.
To encourage our latent variable to capture motion patterns, we applied the KL annealing technique ~\cite{bowman2015generating} during training, in which we gradually increased the weight of KL term from 0 to 1.
For experiments on Aff-Wild only, we applied dropout of ratio 0.8 to both sequence encoder and decoder networks to learn more robust features.

We used Prediction LSTM~\cite{villegas2017learning} as a deterministic baseline. Similar model has been used in previous work for learning dynamics of human motion ~\cite{fragkiadaki2015recurrent,chao2017forecasting}.
We implemented the vanilla VAE model~\cite{mohammad2018stochastic} as our stochastic baseline. 
Similar model has been utilized in \cite{xue2016visual,walker2016uncertain,walker2017pose} for stochastic flow prediction from a single image.
During training, we used $\mathcal{L}_1$ distance as the reconstruction term. 
We conducted extensive hyper-parameter search for vanilla VAE and our MT-VAE variants by enumerating smoothing window $K \in [0, 4, 8, 12, 16]$, motion ratio $\lambda_\text{motion} \in [0, 1, 5, 10, 20]$, cycle loss ratio $\lambda_\text{cycle} \in [0, 1, 5, 10, 20]$.
All models achieve the best performance with $K=8$ and $\lambda_\text{cycle}=5$.
Specifically, the best-performing MT-VAE (add) takes the hyper-parameter $\lambda_\text{motion} = 5$, while all other models take the hyper-parameter $\lambda_\text{motion} = 20$. 

Please visit the website for more visualizations: {\color{blue}\underline{\href{https://goo.gl/2Q69Ym}{https://goo.gl/2Q69Ym}}}.

\cutsubsectionup
\subsection{Multimodal Motion Generation}
\label{sec:exp_seqgen}
\cutsubsectiondown

\begin{table}[t]
\small
\centering
\caption{
Quantitative evaluations for multimodal motion generation.
We compare against two simple data-driven baselines for quantitative comparison:
\textit{Last-step Motion} that recursively applies the motion (velocity only) from the last step observed;
\textit{Sequence Motion} that recursively adds the average sequence velocity from the observed frames.
}
\begin{subtable}{1.0\textwidth}

\caption{Results on Aff-Wild with facial expression coefficients.}
\centering
\begin{tabular}{l||c|c||c|c||c}
\hline
\multirow{2}{*}{Method $/$ Metric}  & \multicolumn{2}{c||}{R-MSE $\downarrow$ ($\times 10^{-1}$)} & \multicolumn{2}{c||}{S-MSE $\downarrow$ ($\times 10^{-1}$)}  & \multirow{2}{*}{Test CLL $\uparrow$ ($\times 10^3$)} \tabularnewline
\cline{2-5} 
 & \texttt{train} & \texttt{test} & \texttt{train} & \texttt{test} & \\
\hline
Last-step Motion & --- & --- & 63.8 $\pm$ 1.31 & 74.7 $\pm$ 5.59 & 0.719 $\pm$ 0.077\\
Sequence Motion & --- & --- & 18.4 $\pm$  0.25 & 19.1 $\pm$ 1.02 & 1.335 $\pm$ 0.057\\
Prediction LSTM~\cite{villegas2017learning} & --- & --- &  1.53 $\pm$ 0.01 & 3.03 $\pm$ 0.06 & 2.232 $\pm$ 0.003 \\
Vanilla VAE~\cite{mohammad2018stochastic} & 0.32 $\pm$ 0.00 & 1.28 $\pm$ 0.02 & 0.79 $\pm$ 0.00 & 1.79 $\pm$ 0.03 & 2.749 $\pm$ 0.012\\
\hline
Our MT-VAE (concat) & 0.22 $\pm$ 0.00 & 0.73 $\pm$ 0.01 & 1.04 $\pm$ 0.00 & 1.76 $\pm$ 0.03 &  2.817 $\pm$ 0.023\\
Our MT-VAE (add) & 0.20 $\pm$ 0.00 & \textbf{0.47} $\pm$ 0.01 & 1.02 $\pm$ 0.00 & \textbf{1.54} $\pm$ 0.04 & \textbf{3.147} $\pm$ 0.018\\
\hline
\end{tabular}
\end{subtable}
\begin{subtable}{1.0\textwidth}
\caption{Results on Human3.6M with 2D joints.}
\centering
\begin{tabular}{l||c|c||c|c||c}
\hline
\multirow{2}{*}{Method $/$ Metric}  & \multicolumn{2}{c||}{R-MSE $\downarrow$} & \multicolumn{2}{c||}{S-MSE $\downarrow$} & \multirow{2}{*}{Test CLL $\uparrow$ ($\times 10^4$)} \tabularnewline
\cline{2-5} 
 & \texttt{train} & \texttt{test} & \texttt{train} & \texttt{test} & \\
\hline
Last-step Motion & --- & --- & 35.2 $\pm$ 0.49 & 32.1 $\pm$ 0.80 & 0.390 $\pm$ 0.004\\
Sequence Motion & --- & --- & 37.8 $\pm$ 0.49 & 35.2 $\pm$ 0.73 & 0.406 $\pm$ 0.003\\
Prediction LSTM~\cite{villegas2017learning} & --- & --- & 1.69 $\pm$ 0.02 & 11.2 $\pm$ 0.17 & 0.602 $\pm$ 0.002\\
Vanilla VAE~\cite{mohammad2018stochastic} & 0.36 $\pm$ 0.00 & 1.05 $\pm$ 0.02 & 3.18 $\pm$ 0.02 & 3.88 $\pm$ 0.05 & 0.993 $\pm$ 0.011\\
\hline
Our MT-VAE (concat) & 0.36 $\pm$ 0.00 & 0.97 $\pm$ 0.02 & 2.26 $\pm$ 0.03 & \textbf{2.84} $\pm$ 0.05 & 1.033 $\pm$ 0.010\\
Our MT-VAE (add) & 0.25 $\pm$ 0.00 & \textbf{0.75} $\pm$ 0.01 & 2.37 $\pm$ 0.02 & \textbf{2.87} $\pm$ 0.05 & \textbf{1.141} $\pm$ 0.009\\
\hline
\end{tabular}
\end{subtable}
\label{tab:table_seqgen}
\vspace*{-0.2in}
\end{table}

We evaluate our model's capacity to generate diverse and plausible future motion patterns for a given sequence on the Aff-Wild and Human3.6M test sets.
Given sequence $S_A$ as initialization, we generated multiple motion trajectories in the future using our proposed sampling and generation process.
For the Prediction LSTM model, we only sample one motion trajectory in the future since the predicted future is deterministic. 

\paragraph{Quantitative Evaluations.}
\vspace*{-0.1in}
We evaluate our model and baselines quantitatively using the minimum squared error metric and conditional log-likelihood metric, which have been used in evaluating conditional generative models \cite{sohn2015learning,walker2016uncertain,yan2016attribute2image,mohammad2018stochastic}.
As defined in Eq.~\ref{eqn:method-RMSE},
\textit{Reconstruction} minimum squared error (or R-MSE) measures the squared error of the closest reconstruction to ground-truth when sampling latent variables from the recognition model. This is a measure of the quality of reconstruction given both current and future sequences. 
As defined in Eq.~\ref{eqn:method-SMSE},
\textit{Sampling} minimum squared error (or S-MSE) measures the squared error of the closest sample to ground-truth when sampling latent variables from prior. 
This is a measure of how close our samples are to the reference future sequences.
\begin{align}
\text{R-MSE} &= \min_{1\leq k \leq K} \| S_B - S^*_B(z^{(k)})\|^2 \text{, where } z^{(k)} \sim q_\phi(z|S_A, S_B).\label{eqn:method-RMSE}\\
\text{S-MSE} &= \min_{1\leq k \leq K} {\|S_B - S^*_B(z^{(k)})\|}^2 \text{, where } z^{(k)} \sim p_\theta(z).\label{eqn:method-SMSE}
\end{align}
In terms of generation diversity and quality, a good generative model is expected to achieve low R-MSE and S-MSE values, given sufficient number of samples.
Note that \textit{posterior collapse} issue is usually featured by low S-MSE but high R-MSE, as latent $z$ sampled from the recognition model is being ignored to some extent.
In addition, we measure the test conditional log-likelihood of the ground-truth sequences under our model via Parzen window estimation (with a bandwidth determined based on the validation set).
We believe that Parzen window estimation is a reasonable approach for our setting as the dimensionality of data (sequence of keypoints) is not too high (unlike in the case of high-resolution videos).
For each example, we used $50$ samples to compute R-MSE metric, and $500$ samples to compute S-MSE and conditional log-likelihood metrics.
On Aff-Wild, we evaluate the models on 32-step expression coefficients prediction (29 $\times$ 32 = 928 dimensions in total).
On Human3.6M, we evaluate the models on 64-step 2D joints prediction (64 $\times$ 64 = 4096 dimensions in total).
Please note that such measurements are approximate, as we do not evaluate the model performance for every sub-sequence (e.g., essentially, every frame can serve as a starting point).
Instead, we repeat the evaluations every 16 frames on Aff-Wild dataset and every 100 frames on Human3.6M dataset.

As we see in Table~\ref{tab:table_seqgen}, data-driven approaches that simply repeat the motion computed from last-step velocity or averaged over the observed sequence performed poorly on both datasets.
In contrast, the Prediction LSTM~\cite{villegas2017learning} baseline greatly reduces the S-MSE metric compared to simple data-driven approaches, due to the deep sequence encoder and decoder architecture in modeling more complex motion dynamics through time.
Among all three models using latent variables, our MT-VAE (add) model achieve the best quantitative performance.
Compared to MT-VAE (concat) that adopts vector concatenation, our additive version achieves lower reconstruction error with similar sampling eror. 
This suggests that the MT-VAE (add) model is able to regularize the learning of motion transformation further.

\begin{figure*}[!h]
\centering
\includegraphics[width=0.99\linewidth]{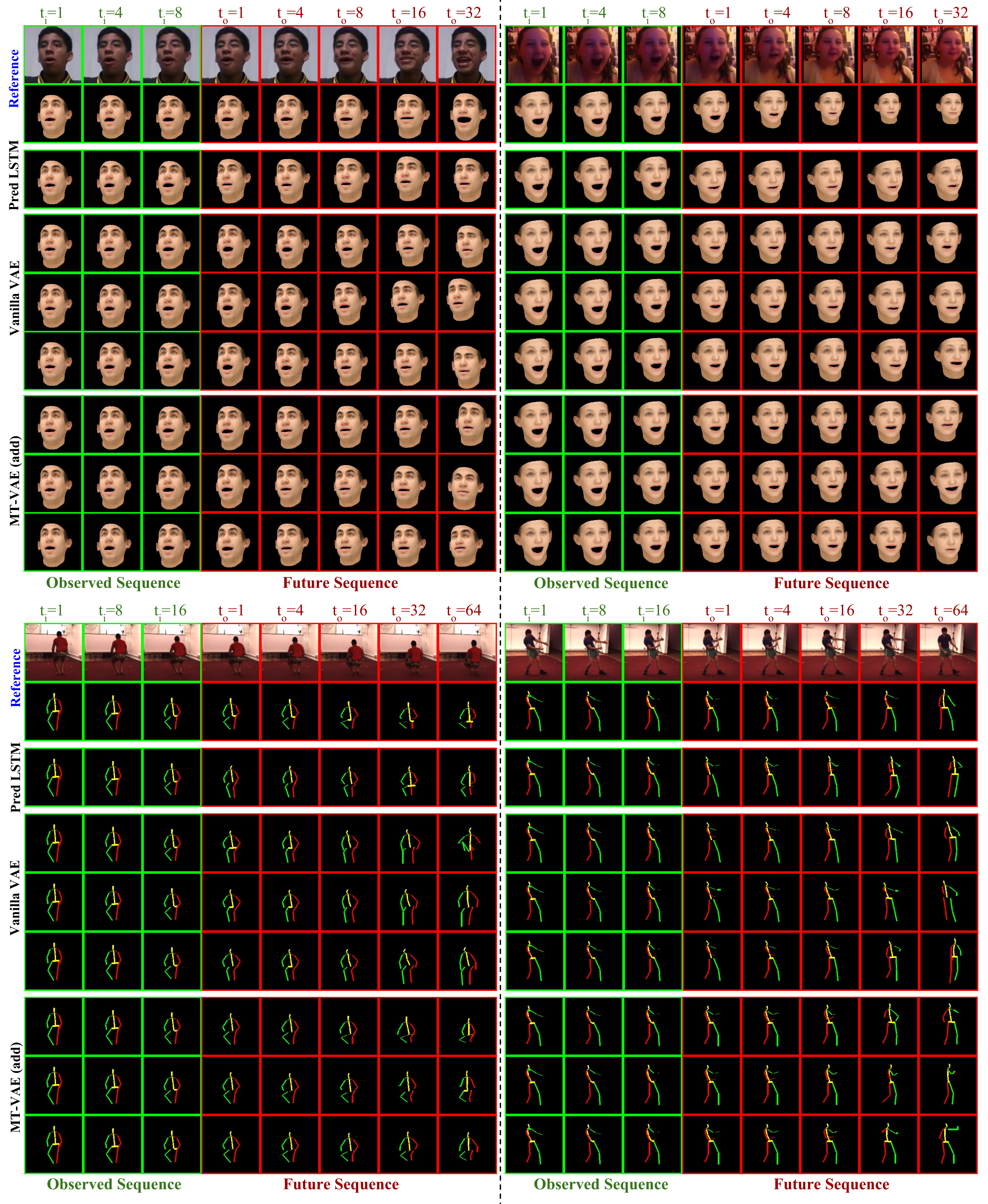}
\caption{Multimodal Sequence Generation. 
Given an input sequence (green boundary), we generate future sequences (red boundary).
We predict 32 frames given 8 frames for face motion, and 64 frames given 16 frames for human body motion.
Given the initial frames as condition, we demonstrate (top to bottom) the ground truth sequence, Prediction LSTM, Vanilla VAE, and our MT-VAE model.
Overall, our model produces (1) diverse and structured motion patterns and (2) more natural transitions from the last frame observed to the first frame generated (See the subtle mouth shape and scale change from the last observed frame to the first generated one).
}
\vspace*{-0.2in}
\label{figure:figure_seqgen}
\end{figure*}

\paragraph{Qualitative Results.}
\vspace*{-0.1in}
We provide qualitative side-by-side comparisons across different models in Figure~\ref{figure:figure_seqgen}.
For Aff-Wild, we render 3D face models using the generated expression-pose parameters along with the original identity parameters.
For Human3.6M, we directly visualize the generated 2D keypoints.
As shown in the generated sequences, our MT-VAE model is able to generate multiple diverse and plausible sequences in the future.
In comparison, the sequences generated by Vanilla VAE are less realistic.
For example, given a sitting down motion (lower-left part in Fig.~\ref{figure:figure_seqgen}) as initialization, the vanilla model fails to predict the motion trend (sitting down), while creating some artifacts (e.g., scale change) in the future prediction.
Also note that MT-VAE produces more natural transitions from the last observed frame to the first generated one (see mouth shapes in the face motion examples and distances between two legs in full-body examples).
This demonstrates that MT-VAE learns a more robust and structure-preserving representation of motion sequences compared to other baselines.

\paragraph{Crowd-sourced Human Evaluations.}

We conducted crowd-sourced human evaluations via Amazon Mechanical Turk (AMT) on 50 videos (10 Turkers per video) from Human3.6M dataset.
This evaluation presents the past action, and 5 generated future actions for each method to a human evaluator and asks the person to select the most (1) realistic and (2) diverse results.
In this evaluation, we also added comparisons to a recently published work~\cite{denton2018stochastic} on stochastic video prediction, which we refer to as SVG.
Table~\ref{tab:table_h36m_AMT} presents the percentage of users who selected each method for each task. 
The Prediction LSTM produces the most realistic but the least diverse result;
Babaeizadeh et al.~\cite{mohammad2018stochastic} produces the most diverse but the least realistic result; Our MT-VAE model (we use the additive variant here) achieves a good balance between 
realism and diversity.

\begin{table}[t]
\small
\caption{
Crowd-sourced Human Evaluations on Human3.6M. *We did not include Prediction LSTM for the diversity evaluation, as it makes deterministic prediction.
}
\centering
\begin{tabular}{l||c|c|c||c}
\hline
Metric  & Vanilla VAE~\cite{mohammad2018stochastic} & SVG~\cite{denton2018stochastic} & Our MT-VAE (add) & Pred LSTM~\cite{villegas2017learning}\\
\hline
Realism (\%)  & 19.2 & 23.8 & 26.4 & 30.6\\
\hline
Diversity (\%)  & 51.6 & 22.3 & 26.1 & ${0.0}^*$\\
\hline
\end{tabular}
\label{tab:table_h36m_AMT}
\vspace*{-0.1in}
\end{table}

\begin{table}[t]
\small
\caption{
Ablation Study on Different variants of MT-VAE (add) model: We evaluate models trained without motion coherence objective, without cycle consistency objective, and the model with context-free latent decoder.
}
\centering
\begin{tabular}{l||c|c|c}
\hline
{Method $/$ Metric} & {R-MSE (test) $\downarrow$} & {S-MSE (test) $\downarrow$} & {Test CLL $\uparrow$ ($\times 10^4$)}\\
\hline
MT-VAE (add) & 0.75 $\pm$ 0.01 & 2.87 $\pm$ 0.05 & 1.141 $\pm$ 0.009\\
\hline
MT-VAE (add) w/o Motion Coherence & 1.01 $\pm$ 0.02 & 2.93 $\pm$ 0.04 & 1.012 $\pm$ 0.014\\
\hline
MT-VAE (add) w/o Cycle Consistency & 1.18 $\pm$ 0.03 & 2.71 $\pm$ 0.05 & 0.927 $\pm$ 0.019\\
\hline 
MT-VAE (add) Context-free Decoder & 0.31 $\pm$ 0.05 & 4.05 $\pm$ 0.05 & 1.299 $\pm$ 0.007\\
\hline
\end{tabular}
\label{tab:table_h36m_ablation}
\vspace*{-0.1in}
\end{table}

\paragraph{Ablation Study.}
\vspace*{-0.1in}

We analyze variations of our MT-VAE (add) models on Human3.6M.
As we see in Table~\ref{tab:table_h36m_ablation}, removing the cycle consistency or motion coherence  results in a drop in reconstruction performance.
This shows that cycle consistency and motion coherence encourage the motion feature to preserve motion structure and hence be more discriminative in nature.
We also evaluate a \textit{context-free} version of the MT-VAE (add) model, where the the transformation vector $\mathcal{T}^*$ is not conditioned on input feature $e_A$. This version  produces poor S-MSE value since it is challenging for the additive latent decoder to hallucinate transformation vector $\mathcal{T}^*$ solely from latent variable $z$.

\cutsubsectionup
\subsection{Analogy-based Motion Transfer}
\cutsubsectiondown

We evaluate our model on an additional task of \emph{transfer by analogy}. 
In this analogy-making experiment, we are given three motion sequences A, B (which is the subsequent motion of A), and C (which is a different motion sequence).
The objective is to recognize the \emph{transition} from A to B and transfer it to C. 
This experiment can demonstrate whether our learned latent space models the mode transition across motion sequences.
Moreover, this task has numerous graphics applications like transferring expressions and their styles, video dubbing, gait style transfer, and video-driven animation~\cite{thies2016face2face}.  

\begin{figure*}[t]
\centering
\includegraphics[width=0.99\linewidth]{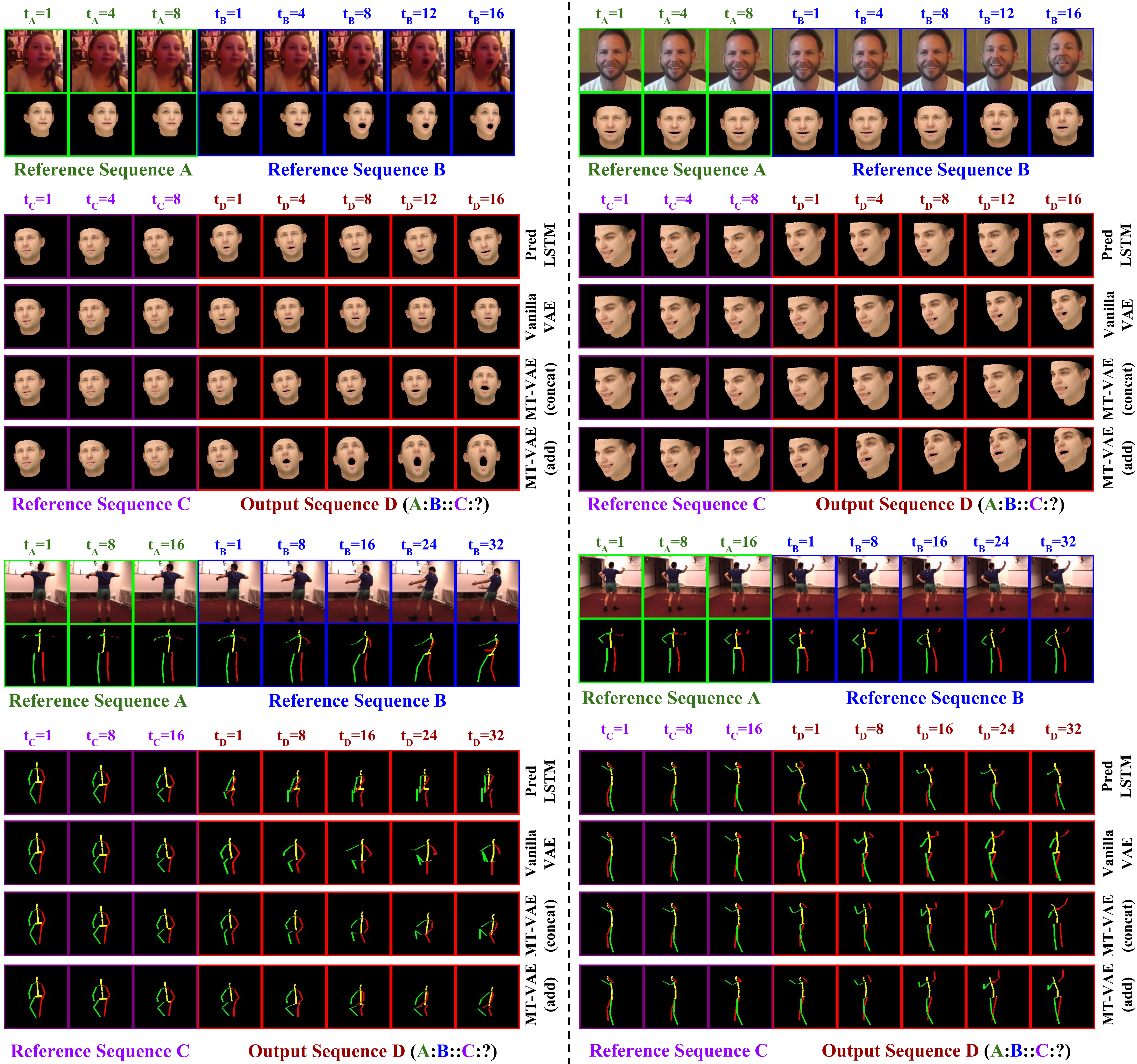}
\caption{Analogy-based motion transfer.
Given three motion sequences A, B, and C from test set, the objective is to extract the motion mode transition from A to B and then apply it to animate the future starting from sequence C.
For fair comparison, we set the encoder Gaussian distribution parameter $\sigma$ to zero during evaluation.
}
\label{figure:figure_seqanalogy}
\vspace*{-0.2in}
\end{figure*}

In this experiment, we compare Prediction LSTM, Vanilla VAE, and our MT-VAE variants.
For the stochastic models, we compute the latent variable $z$ from motion sequence A and B via the latent encoder, i.e., $z = h_{\mathcal{T} \rightarrow z}(e_B - e_A)$, and then decode using motion sequence C as $e^*_D = h_{z \rightarrow \mathcal{T}}(z, e_C)$.
For Prediction LSTM model, we directly performed the analogy-making in the feature space $e^*_D = e_B - e_A + e_C$ since there is no notion of a latent space in that model.
As shown in Figure~\ref{figure:figure_seqanalogy}, our MT-VAE model is able to combine the transformation learned from A to B transitions with the structure in sequence C. The other baselines failed at either adapting the mode transition from A to B or preserving the structure in C.
The analogy-based motion transfer task is significantly more challenging than motion generation, since the combination of three reference motion sequences A, B, and C may never appear in the training data.
Yet, our model is able to synthesize realistic motions.
Please note that motion modes may not explicitly correspond to semantic motions, as we learn the motion transformation in an unsupervised manner.

\cutsubsectionup
\subsection{Towards Multimodal Hierarchical Video Generation}
\cutsubsectiondown

\begin{figure*}[t]
\centering
\includegraphics[width=0.99\linewidth]{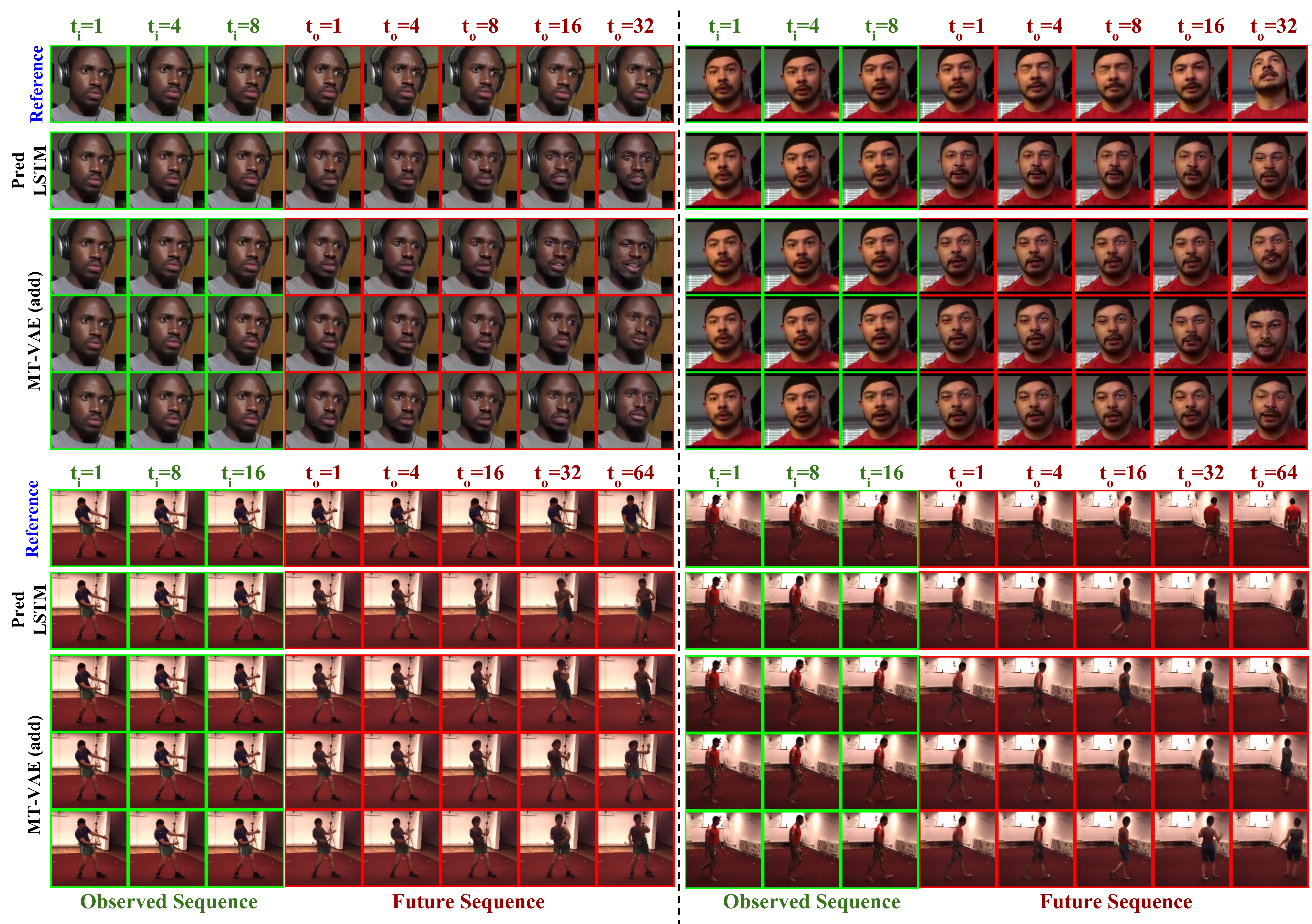}
\caption{Multimodal Hierarchical video generation. Top rows: Face video generation results from 8 observed frames.
Bottom rows: Human video generation results from 16 observed frames.
}
\label{figure:figure_videogen}
\vspace*{-0.2in}
\end{figure*}

As an application, we showcase that our multimodal motion generation framework can be directly used for generating diverse and realistic pixel-level video frames in the future.
We trained the keypoint-conditioned image generation model~\cite{villegas2017learning} that takes both previous image frame A and predicted motion structure B (e.g., rendered face or human joints) as input and hallucinates image C by combining the image content adapted from A but with motion adapted from B.
In Figure~\ref{figure:figure_videogen}, we show a comparison of video generated in a deterministic way by Prediction LSTM (i.e., single future), and in a stochastic way driven by the predicted motion sequence (i.e., multiple futures) from our MT-VAE (add) model.
We use our generated motion sequences for performing video generation experiments on the Aff-Wild (with 8 input frames observed) and Human3.6M (with 16 input frames observed).

\cutsectionup
\section{Conclusions}
\cutsectiondown

Our goal in this work is to learn a conditional generative model for human motions. 
This is an extremely challenging problem in the general case and can require significant amount of training data to generate realistic results.
Our work demonstrates that this can be accomplished with minimal supervision by enforcing a strong structure on the problem.
In particular, we model long-term human dynamics as a set of motion modes with transitions between them, and construct a novel network architecture that strongly regularizes this space and allows for stochastic sampling.
We have demonstrated that this same idea can be used to model both facial and full body motion, independent of the representation used (i.e., shape parameters, keypoints).

\clearpage
\subsubsection*{Acknowledgements.}
\cutsectiondown
We thank Zhixin Shu and Haoxiang Li for their assistance with face tracking and fitting codebase. 
We thank Yuting Zhang, Seunghoon Hong, and Lajanugen Logeswaran for helpful comments and discussions.
This work was supported in part by Adobe Research Fellowship to X. Yan, a gift from Adobe, ONR N00014-13-1-0762, and NSF CAREER IIS-1453651.

{\small
\bibliographystyle{splncs}
\bibliography{references}
}

\end{document}